\documentclass{article} 
\usepackage{nips15submit_e,times}
\usepackage{hyperref}
\usepackage{url}
\usepackage{authblk}
\usepackage{amsmath,graphicx,booktabs}
\usepackage{epstopdf,amssymb}
\usepackage{multirow}
\usepackage{fixltx2e}
\usepackage{subfigure}
\usepackage{float}

\title{\textbf{A Strategy of MR Brain Tissue Images$^{'}$ Suggestive Annotation Based on Modified U-Net}} 

\author{Yang Deng$^{1,2}$, Yao Sun$^{1,2}$, Yongpei Zhu$^{1,2}$,  Mingwang Zhu$^{3}$,Wei Han$^{4*}$,Kehong Yuan$^{1*}$\\
	$^{1}$Graduate School at Shenzhen, Tsinghua University, Shenzhen 518055, China.\\
	$^{2}$Department of Biomedical Engineering, Tsinghua University, Beijing 100084, China.\\
	$^{3}$Beijing Sanbo Brain Hospital, Beijing 100825, China.\\
	$^{4}$The First Affiliated Hospital of Harbin Medical University, Harbin 150001, China
	*Corresponding author: Han Wei(e-mail:dr.hanwei@foxmail.com) and Kehong Yuan (e-mail:yuankh@sz.tsinghua.edu.cn)
}

\nipsfinalcopy 

\begin{document}

\maketitle

\begin{abstract}
Accurate segmentation of MR brain tissue is a crucial step for diagnosis, surgical planning, and treatment of brain abnormalities. However, it is a time-consuming task to be performed by medical experts. So, automatic and reliable segmentation methods are required. How to choose appropriate training dataset from limited labeled dataset rather than the whole also has great significance in saving training time. In addition, medical data labeled is too rare and expensive to obtain extensively, so choosing appropriate unlabeled dataset instead of all the datasets to annotate, which can attain at least same performance, is also very meaningful. To solve the problem above, we design an automatic segmentation method based on U-shaped deep convolutional network and obtain excellent result with average DSC metric of 0.8610, 0.9131, 0.9003 for Cerebrospinal Fluid (CSF), Gray Matter (GM) and White Matter (WM) respectively on the well-known IBSR18 dataset. We use bootstrapping algorithm for selecting the most effective training data and get more state-of-the-art segmentation performance by using only 50\% of training data. Moreover, we propose a strategy of MR brain tissue images’ suggestive annotation for unlabeled medical data based on the modified U-net. The proposed method performs fast and can be used in clinical.\\

\end{abstract}

\section{Introduction}
 With the advent of aging society, degenerative diseases of the central nervous system are becoming more common. Degenerative diseases of the central nervous system refer to a group of diseases caused by degeneration of chronic progressive central nervous system. The degeneration and loss of neurons in the brain spinal cord can be seen in pathology. Magnetic resonance (MR) can be more clearly and safer to display the structure of the brain because of its non-invasive, non-radioactive, free selection profile, higher signal to noise ratio, and higher resolution of the soft tissue with smaller density difference, so as to provide more information for the pathological diagnosis of brain diseases. It has become a common method for the examination of brain diseases. The precise segmentation of brain tissue is the first step of the volume and quantitative analysis of the brain. It is of great significance to the diagnosis and treatment of brain diseases, especially the neurodegenerative diseases, and the discovery of many subsequent neurological diseases.\\

The classical segmentation methods are mainly classified as follows: (1) threshold segmentation (2) segmentation based on regional growth (3) segmentation based on the watershed algorithm (4) segmentation based on the statistical algorithms like the mixed Gauss model (GMM) and Markov random field (MRF) (5) segmentation based on Atlas (6) segmentation based on clustering algorithm like K mean (K-means), fuzzy C mean (FCM) (7) a mixed application of the above algorithm \cite{Qin2017Brain} \\	

Since AlexNet \cite{Krizhevsky2012ImageNet} won the ImageNet Challenge in 2012, deep learning attracted the attention of researchers again. Over the last few years, deep learning especially deep convolutional neural networks (CNNs) have emerged as one of the most prominent approaches for image recognition problems in various domains. There are three main ideas for the segmentation of brain MRI image with convolution neural network \cite{Akkus2017Deep}: (1) Patch-Wise CNN Architecture. This is a simple approach to train a CNN algorithm for segmentation. A fixed size patch around each pixel is extracted from a given image, and the model is trained on these patches with labels of the patches’ center pixel classes, such as normal brain and tumor. The disadvantages of this method are huge computation and hard to train (2) Semantic-Wise CNN Architecture \cite{Long2015Fully} \cite{Ronneberger2015U}. This type of architecture makes predictions for each pixel of the whole input image, which the network needs only one forward inference. This kind of architecture includes encoder part that extracts features and decoder part that combines lower level features from the encoder part to form abstract features. The input image is mapped to the segmentation labels in a way that minimizes a loss function. (3) Cascaded CNN Architecture \cite{Dou2016Automatic}. This type of architecture combines two CNN architectures. The first CNN is used to train the model for preliminary prediction and the second CNN is used to further adjust the prediction of the first network.\\	

Zhang et al.\cite{Nie2015FULLY} proposed to use CNN method based on image patches to segment gray matter, white matter and cerebrospinal fluid from multimodal baby MR images, and performed better compared with the traditional methods; Nie et al.\cite{Nie2015FULLY} proposed a semantic level full convolution network segmentation method and got higher dice similarity coefficient (DSC) than Zhang’s method; Moeskops et al. \cite{Moeskops2016Automatic} proposed a multi-scale ( pixels) pach-wise CNN method to segment brain images of infants and young adults; Bao et al. \cite{Bao2015Multi}also proposed a multi-scale patch-wise CNN method together with dynamic random walker with decay region of interest to obtain smooth segmentation of subcortical structures in IBSR (developed by the Centre for Morphometric Analysis at Massachusetts General Hospital-available at https://www.nitrc.org/projects/ibsr to download) and LPBA40 datasets; Chen et al. \cite{Chen2017VoxResNet} proposed deep voxelwise residual networks for volumetric brain segmentation.\\	

Compared to CNN, FCN has no the full connection layer, which makes it possible to input any size of the image, to greatly reduce the training parameters and to improve the speed of training. U-shaped deep convolutional network, as a kind of outstanding FCN, has excellent performance for biomedical image segmentation and almost becomes the preferred method due to the advantage of light, flexible, strong robustness and needing a small amount of data. In this paper, we designed a U-shaped net to segmentation the MRI brain tissue and proposed a strategy of MR brain tissue images’ suggestive annotation based on our model, both of which achieved promising result.\\	

The remainder of this paper is organized as follows. In section 2, we present our method. Experiments and results are detailed in section 3. Finally, the discussion and the main conclusions are presented in section 4 and section 5 separately.\\

\section{Method}
\subsection{U-shape Network Architecture}
Figure 1 illustrates the network architecture we use in this paper. Like the standard U-Net \cite{Ronneberger2015U}, it has an analysis and a synthesis path each with four resolution steps.

\begin{figure}[H]
\begin{center}
\includegraphics[width=1.\linewidth]{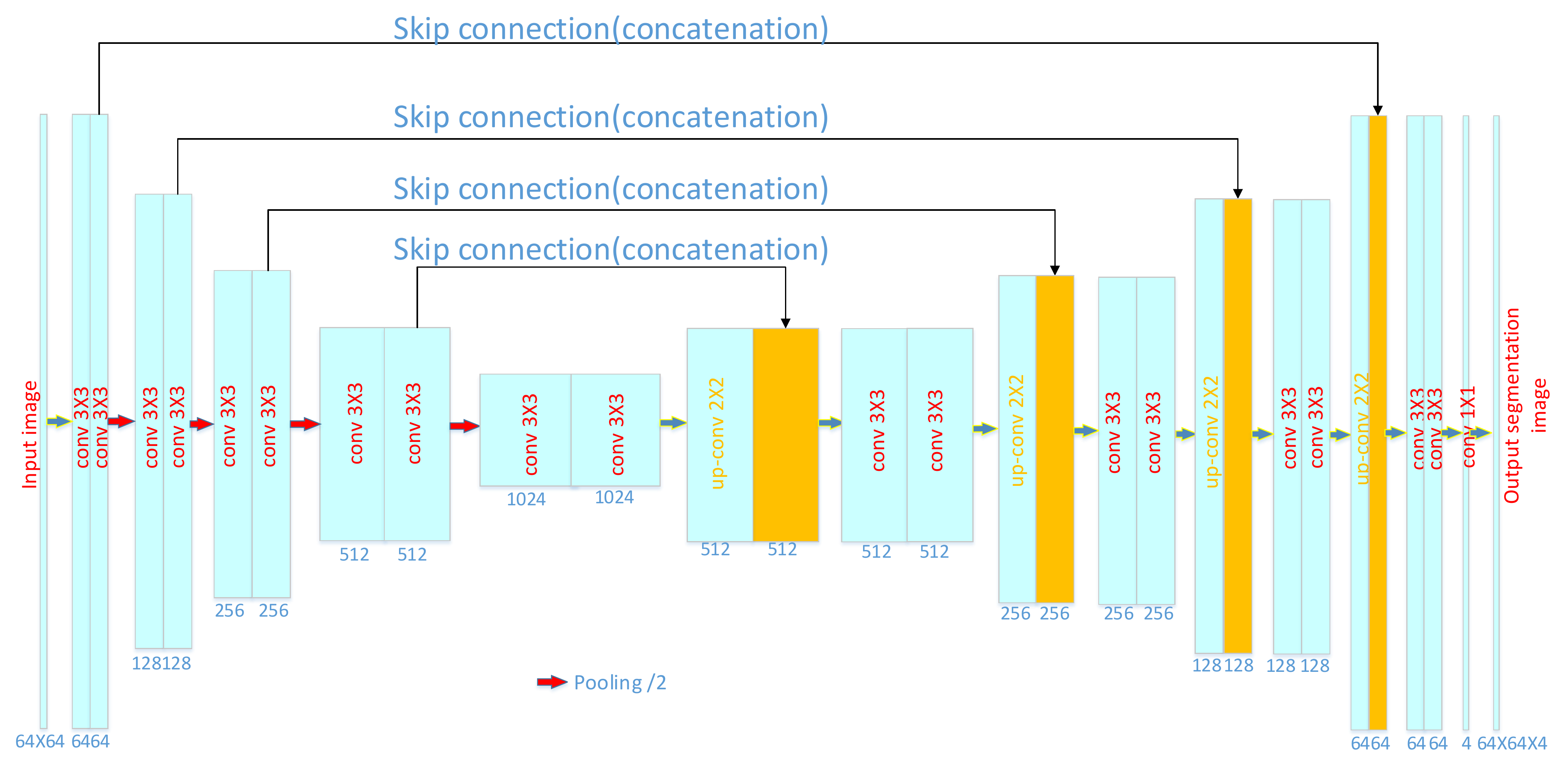}
\caption{The modified U-Net}
\end{center}
\end{figure}

In the analysis path, each layer contains two 3×3 convolutions each followed by a rectified linear unit (ReLu), and then a 2×2 max pooling with strides of 2 for down-sampling. In the synthesis path, each layer consists of an upconvolution of 2×2 by strides of one in each dimension, followed by two 3×3 convolutions each followed by a ReLu. To keep the same shape after convolution, we use padding. Shortcut connections from layers of equal resolution in the analysis path provide the essential high-resolution features to the synthesis path \cite{?20163D}. In the last layer, a 1×1 convolution reduces the subject of output channels to the subject of labels which is 4 in our case. The architecture has 31030788 parameters in total.\\	

Like suggested in \cite{Szegedy2016Rethinking} we avoid bottlenecks by doubling the subject of channels already before max pooling. We also adopt this scheme in the synthesis path. The input size to the network is 64x64 and the output is 64x64x4, where the four channel corresponds to the background, CSF, GM, and WM.\\

Different from the purely U-net\cite{Ronneberger2015U}, our network can segment CSF, GM and WM three tissues at once because we use our own loss function which will be detailed in section 3. At the same time, the shape of output is the same as input owing to the use of padding.\\

\subsection{Bootstrapping algorithm for selecting the most effective training data} From Lin Yang’s paper \cite{Yang2017Suggestive}, we know that the quantity and quality of train dataset is very important for the model. Lin Yang’s paper indicates that state-of-the-art segmentation performance can be achieved by using only 50\% of training data. Inspired by this result, we use bootstrapping algorithm to pick out the most effective 50\% training data on IBSR18 dataset. Bootstrapping \cite{Efron1993An} is a standard way for evaluating the uncertainty of learning models. Its basic idea is to train a set of models while restricting each of them to use a subset of the training data (generated by sampling with replacement) and calculate the variance (disagreement) among these models. The measure of variance we used in this paper is dice similarity coefficient and lower dice similarity coefficient means higher variance. \\	

To ensure to choose the training data randomly, we focus on the data of subject 01,02,03,04,05,10,11,12,13,14 and what we want to do is to choose the more valuable data in this ten datasets, which make the model more excellent. The evaluation standard to determine whether the selected data is more effective than others is the average dice similarity coefficient of remaining 8 sets of testing data.\\	

\begin{figure}[H]
\vspace{-3cm}
\centering
	\includegraphics[width=1.\linewidth]{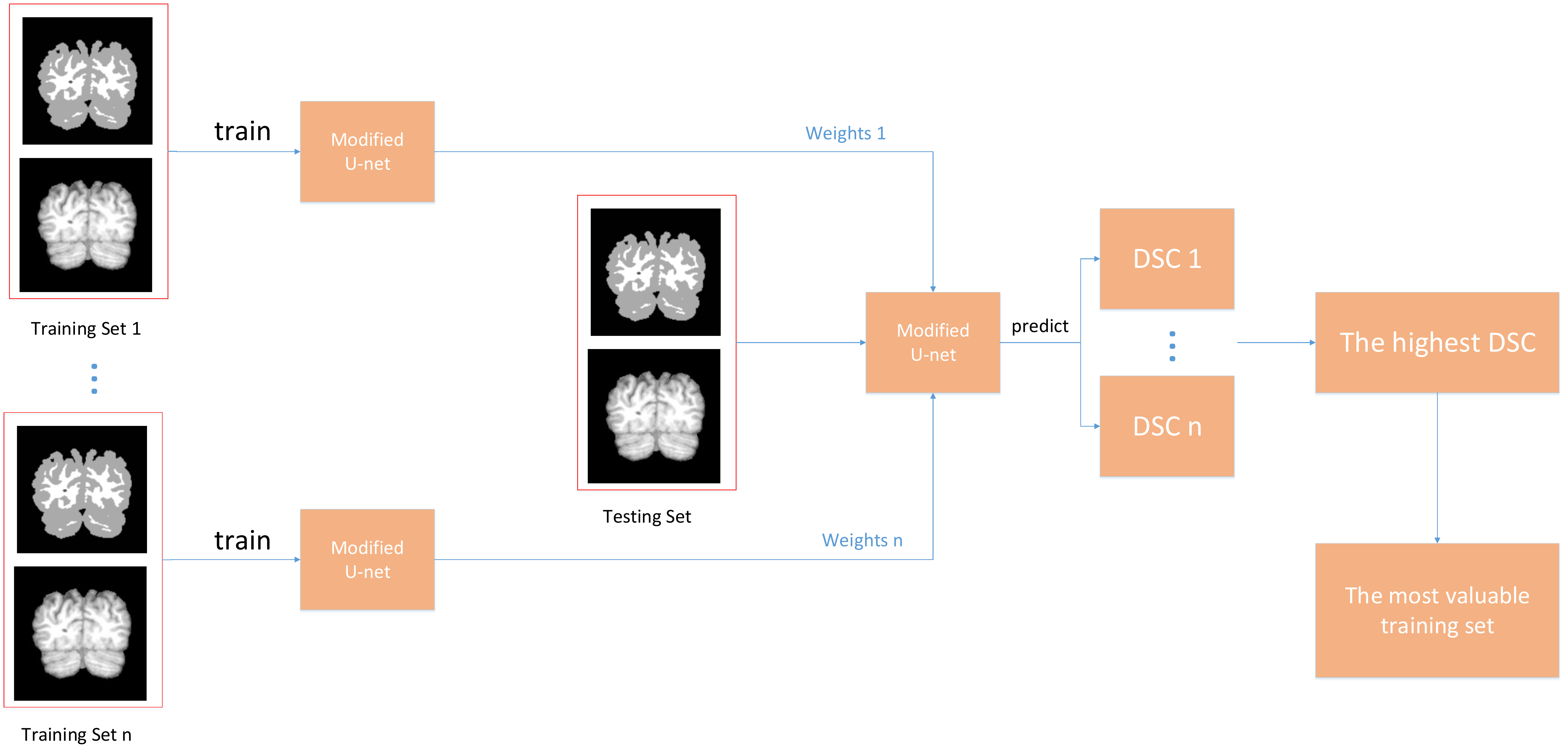}
\caption{The process to find the more valuable data from training dataset}
\end{figure}

\subsection{The Strategy of MR Brain Tissue Images’ Suggestive Annotation Based on Modified U-Net}
Figure 3 shows the process to find the more valuable data from unlabeled dataset. We used the training data selected in section 2.2(dataset subject 3, 10, 11, 12, 13) to train our segmentation model. When the model was trained, we used it to predict the unlabeled data (in this paper, we abandoned the labels of dataset subject 6, 7, 8, 9, 15, 16, 17, 18 and regarded them as unlabeled dataset) and got the predicted label. Then, we combined this label with the preprocessed unlabeled data as new training sets and separately added them into the original training set one at a time to train the model again. We kept repeating this process with fixed epochs and got the unlabeled data’s model. Finally, we got 8 models for 8 unlabeled testing sets. Next, we used the 8 new trained model to predict the testing set (dataset subject 1, 2, 4, 5, 14) and attained 8 average DSC of testing set for the 8 model. Finally, we sorted the DSC according to their numeric value. The lower DSC meant the higher variance and should be label.

\begin{figure}[H]
\begin{center}
\includegraphics[width=1.\linewidth]{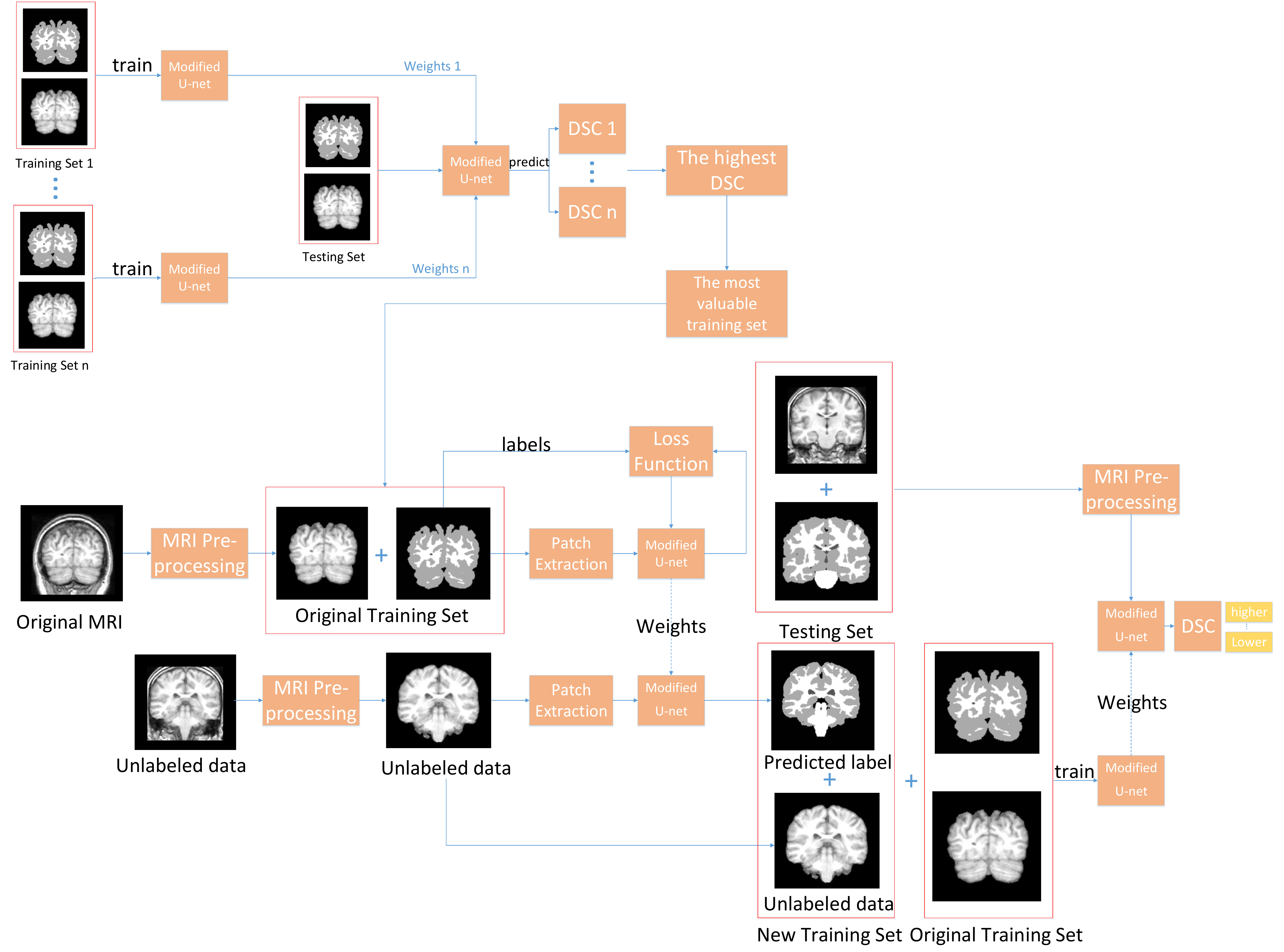}
\caption{The process to find the more valuable data from unlabeled dataset}
\end{center}
\end{figure}

\section{Rusults}
\subsection{Dataset and pre-processing}
We validated our method on the well-known IBSR18 dataset\footnote{\url{https://www.nitrc.org/frs/?group_id=48}}, which is one of the standard datasets for tissue quantification and segmentation evaluation. The dataset consists of 18 MRI volumes and the corresponding ground truth (GT) is provided.\\

Typical preprocessing steps for structural brain MRI include the following key steps \cite{Akkus2017Deep}: registration, skull stripping, bias field correction, intensity normalization and noise reduction. With advent of deep learning techniques, some of the preprocessing steps became less critical for the final segmentation performance \cite{Akkus2017Deep}. In the given dataset, skull-stripping and bias field correction algorithm were already applied. Thus, we do not need to apply these techniques.\\

\subsection{Segmentation quantitative evaluation and comparison}
In this paper, three metrics are used to evaluate the segmentation result: DSC (Dice's coefficient), HD (Hausdorff distance) and AVD (Absolute Volume Difference). First, the DSC, is the most used metric in the evaluation of medical volume segmentations. In addition to the direct comparison between automatic and ground truth segmentations, it is common to use the DSC to measure reproducibility (repeatability) \cite{Taha2015Metrics}. DSC is computed by:
\begin{equation}
DSC=\frac{2TP}{2TP+FP+FN} 
\end{equation}

Where TP, FP and FN are the subjects of true positive, false positive and false negative predictions for the considered class.	
Second, the distance between crisp volumes (HD) between two finite point sets A and B is defined by:
\begin{equation}
H(A,B)=max(h(A,B),h(B,A))
\end{equation}	

Where h(A,B)is called the directed HD and given by:
\begin{equation}
h(A,B)=max(a\in A)min(b\in B)||a-b||
\end{equation}
\begin{equation}
h(B,A)=max(b\in B)min(a\in A)||b-a||
\end{equation}
Finally, the AVD is defined by:
\begin{equation}
AVD(A,B)=\frac{||A-B||}{\sum(A)} 
\end{equation}

Where A is ground truth and B is predicted volume of one class.\\

Regarding the evaluation of testing data, we compared our method with several state-of-the-art methods, including Moeskops’ multi-scale ( pixels) patch-wise CNN method\cite{Bao2015Multi}and Chen’s voxel based residual network	\cite{Chen2017VoxResNet}. For all the there methods, we used IBSR subject 01-05 as training set and the remaining 13 subjects as testing set.

\begin{figure}[H]

	\begin{center}
		\includegraphics[width=1.\linewidth]{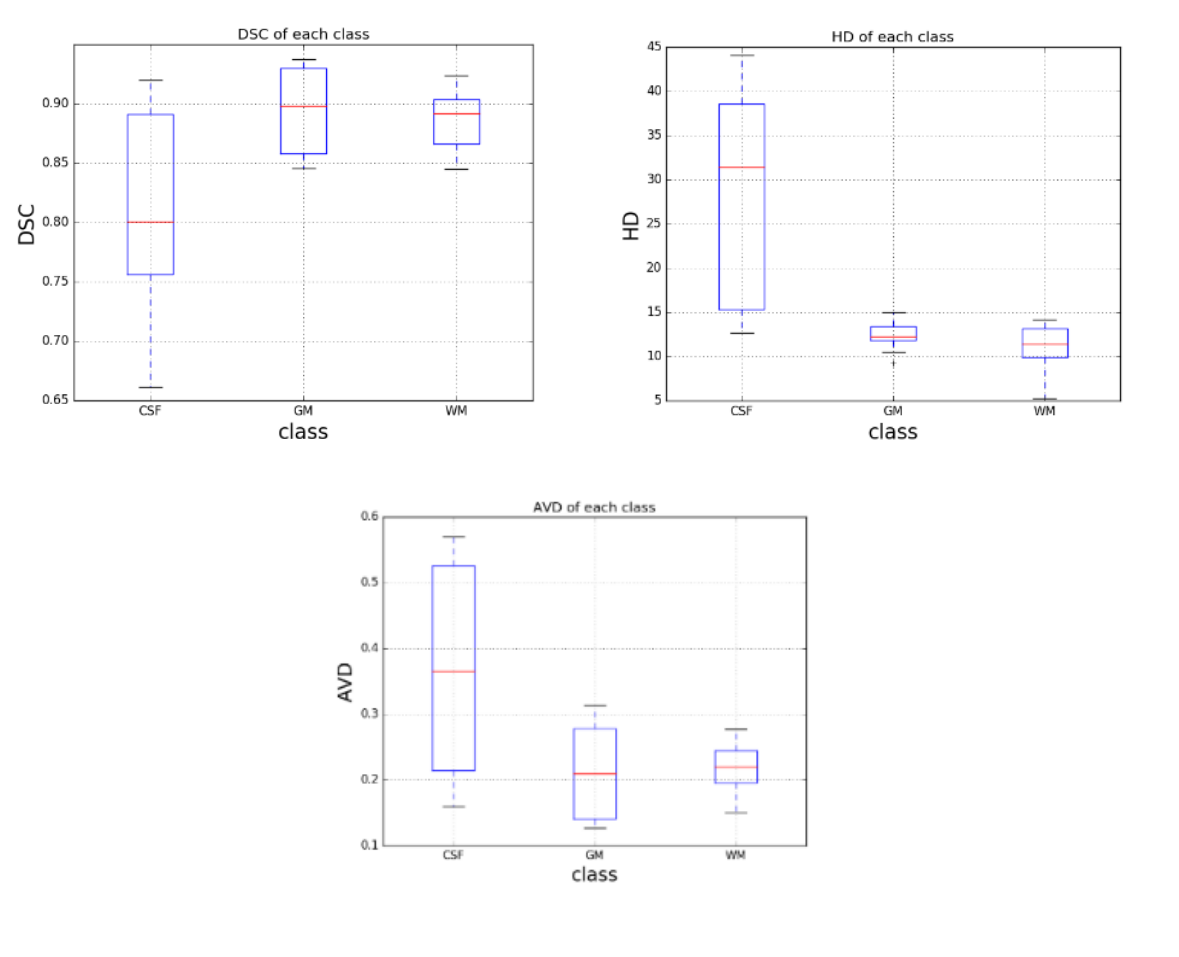}
	\end{center}
	\caption{Boxplot of each metrics with modified U-Net}
\end{figure}

\begin{table}[th]
	\vspace{0em}
	\centering
	\resizebox{1.0\linewidth}{!}{
		\begin{tabular}{ | l | l | l | l | l | l | l | l | l | l | c | }
			\hline
			\multirow{2}{*}{Method} & \multicolumn{3}{c|}{CSF} & \multicolumn{3}{c|}{GM} & \multicolumn{3}{c|}{WM} & \multirow{2}{*}{Time(s)}  \\
			\cline{2-10}
			 & DC & HD & AVD & DC & HD & AVD & DC & HD & AVD &  \\ \hline
			Ours&\textbf{81.16}&\textbf{28.27}&\textbf{37.50}&\textbf{89.46}&12.47&21.11& 88.73 &11.39&\textbf{21.65}&\textbf{40}\\\hline
			VoxResNet~\cite{Chen2017VoxResNet}&79.55&30.39&42.12&88.09& \textbf{12.08}&\textbf{18.15}&\textbf{88.92}&\textbf{9.15}&21.89&100\\\hline
			
			Multi-scale CNN~\cite{Bao2015Multi}&63.01&---&---&80.53&---&---& 82.16&---&---&3500\\\hline
		
		\end{tabular}
				}
		\caption{Results of MR Brain Segmentation using Different Methods (HD: mm)( Moeskops$^{’}$ method is too hard to train and the cost of time is huge, we chose the best result of several experiments here, the metrics (AVD and HD) have been unfortunately lost, so we show the DSC only. ) (DC: \%, HD: mm, AVD: \%).}	
\end{table}

From Table 1, we can see that the modified U-Net achieved quite good performance than others.

%
%
%
%
%

\subsection{The most effective training data}
With bootstrapping algorithm, we got the result of different training sets below.\\	

\begin{table}[H]
	\vspace{-1em}
	\setlength{\belowcaptionskip}{0cm}
	\begin{tabular}{ccccc}
		\toprule
		Training set& Dice of CSF& Dice of GM&Dice of WM&Average dice\\
		\midrule
		1& 0.8062& 0.8886&0.8851&0.8600\\
		2& 0.8296& 0.8961&0.8832&0.8697\\
		3& 0.8525& 0.9114&0.9010&0.8883\\
		\textbf{4}& \textbf{0.8678}&\textbf{0.9185} &\textbf{0.9070}&\textbf{0.8978}\\
		5& 0.8102& 0.8873&0.8772&0.8582\\
		\bottomrule
	\end{tabular}
	\caption{The result of different training sets used on modified U-Net (training set 1 including subject 1,2,4,5,14;training set 2 including subject 1,2,3,4,5;training set 3 including subject 10,11,12,13,14; training set 4 including subject 3,10,11,12,13 and training set 5 including subject 1,2,3,4,5,10,11,12,13,14. Testing set are subject 6 7 8 9 15 16 17 18)}
\end{table}

Fig. 5 is the different results while using different training sets on modified U-Net in table2.

\begin{figure}[H]
	\begin{center}
		\includegraphics[width=1.\linewidth]{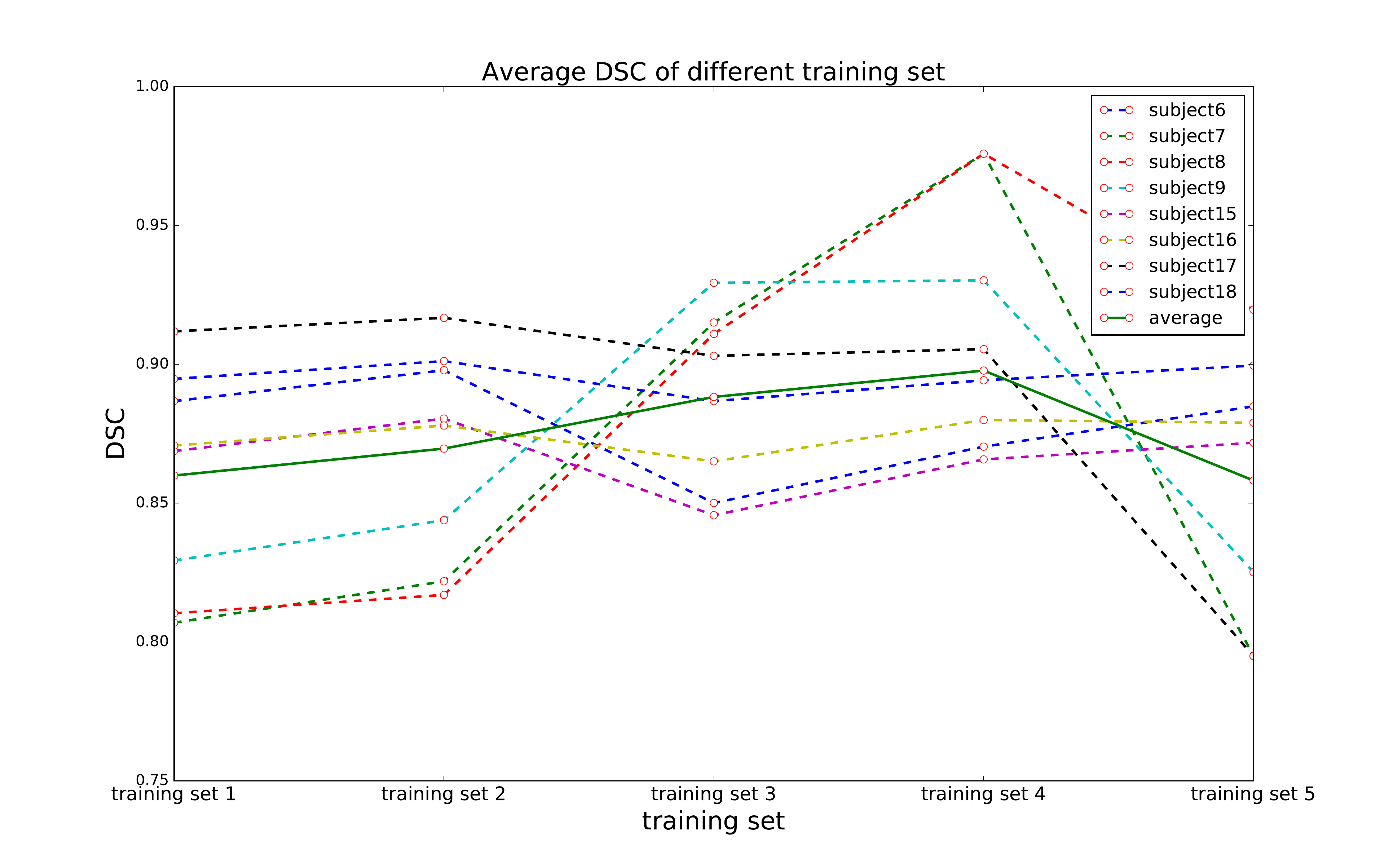}
	\end{center}
	\caption{The different results while using different training sets}
\end{figure}	
From Table 2, we find that different training set has great influence on the result and the model performance is not always better with more training data, even the model will be bad when the training set is too many. So picking out the more valuable training set has great significance for saving training time and even for improving the performance of model.

\subsection{Suggestive Annotation}
We used the most effective training set acquired from section 3.3 to predict the dataset composed of subject 6, 7, 8, 9, 15, 16, 17, 18, which we assumed that they were not labeled. Using the method in section 2.3, we added the predicted label with its preprocessed image to the original training set one at a time to become new training set and predicted testing set (subject 1, 2, 4, 5, 14). Following are the results.

\begin{table}[H]
	\vspace{0em}
	\begin{tabular}{ccccc}
		\toprule
		Training set(subject)& Dice of CSF& Dice of GM&Dice of WM&Average dice\\
		\midrule
		3 10 11 12 13 \textbf{6}& 0.8440& 0.9071&0.8912&0.8807\\
		3 10 11 12 13 \textbf{7}& 0.8465& 0.8988&0.8848&0.8767\\
		3 10 11 12 13 8& 0.8548& 0.9078&0.8919&0.8848\\
		3 10 11 12 13 \textbf{9}& 0.8566& 0.9020&0.8871&0.8819\\
		3 10 11 12 13 15& 0.8650& 0.9085&0.8915&0.8883\\
		3 10 11 12 13 16& 0.8530& 0.9053&0.8913&0.8832\\
		3 10 11 12 13 \textbf{17}& 0.8462& 0.9068&0.8903&0.8811\\
		3 10 11 12 13 18& 0.8578& 0.9111&0.8960&0.8883\\		
		
		\bottomrule
		
	\end{tabular}
	\caption{The result with different training sets on testing set}
\end{table}

From Table 3, we think the subject 7, 6, 17, 9 are the suggestive annotation because their DSCs are lower. To prove whether the data we picked out is more valuable, we used bootstrapping method like section 2.2 because the data we used here (subject 6, 7, 8, 9, 15, 16, 17, 18) have been actually already labeled. Results are below (testing sets here are themselves):	

\begin{table}[H]
	\begin{tabular}{ccccc}
		\toprule
		Training set(subject)& Dice of CSF& Dice of GM&Dice of WM&Average dice\\
		\midrule
		15 18 8 16& 0.7976& 0.8922&0.8824&0.8574\\
		7 8 9 15& 0.8071& 0.8879&0.8808&0.8586\\
		\textbf{7 6 17 9}& 0.8168& 0.8900&0.8790&0.8619\\
		6 7 8 9 15 16 17 18 & 0.7928& 0.8860&0.8746&0.8511\\
		\bottomrule
		
	\end{tabular}
	\caption{The result of proving suggestive annotation}
\end{table}

From Table 4, we know that the dataset we picked out using our strategy are actually worth to be labeled because they get higher DSC.

\subsection{Implementation details and Computation cost}
The project was to find a method which returns the highest average DSC of three tissues, we made our own loss function as following:	

\begin{equation}
L(y,y')=4-\sum_{i=0}^{3}DSC(y_i,y_i')
\end{equation}

where $y_i$ and $y_i'$ are predicted and ground-truth for class i, respectively.

In the stage of segmentation reconstruction, we found the maximum probability among four classes and returned the corresponding label for each voxel rather than finding the optimized threshold.	
\begin{figure}[H]
	\begin{center}
		\includegraphics[width=1.\linewidth]{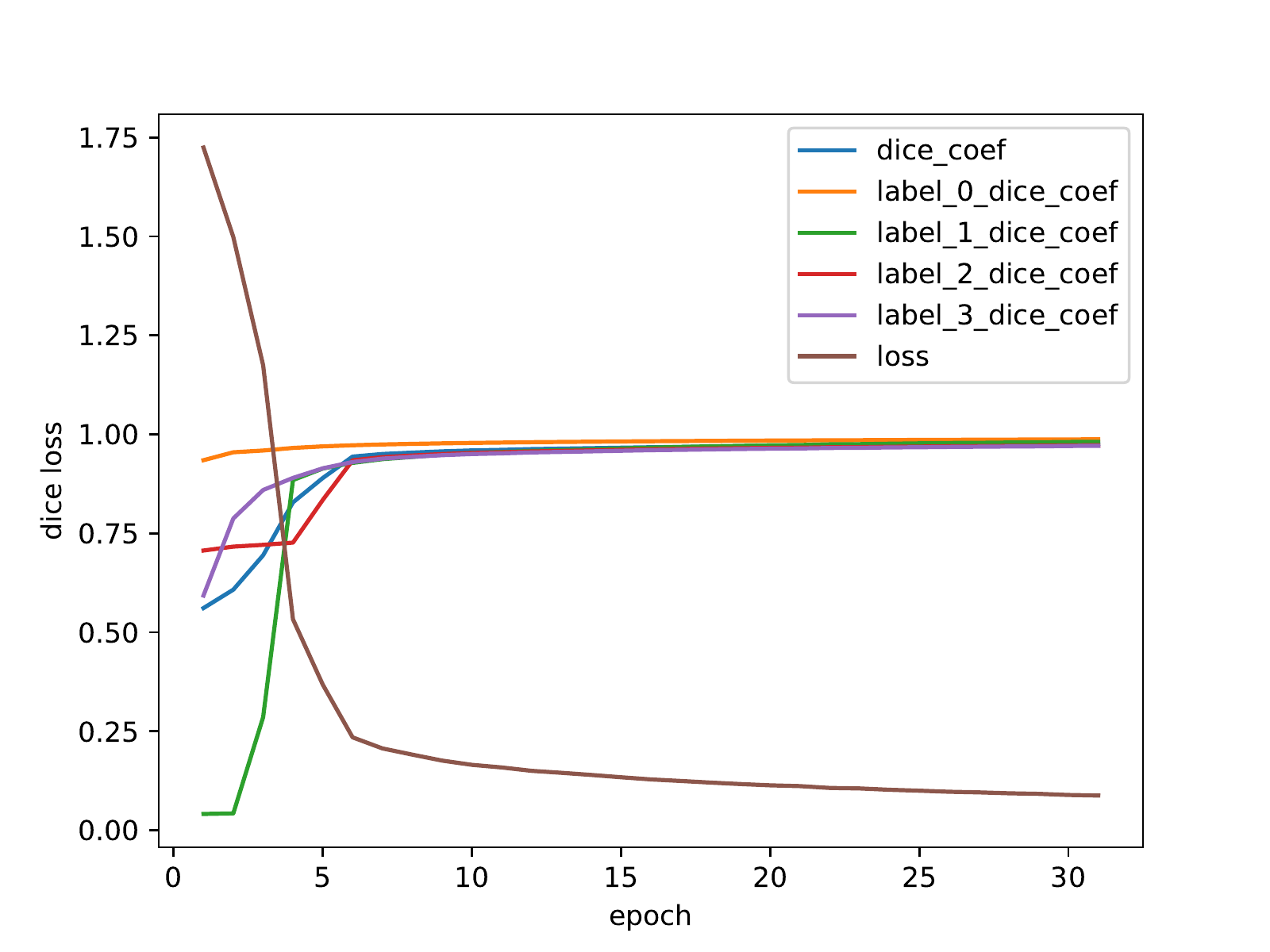}
	\end{center}
	\caption{Dice loss as a function of a training epoch for our proposed models.(label\_0\_dice-coef, label\_1\_dice-coef , label\_2\_dice-coef , label\_3\_dice-coef means the DSC of background, CSF, GM and WM respectively.)}
\end{figure}

The network was trained for 500 epochs on a single NVIDIA TitanX GPU. In order to prevent the network from over-fitting, we applied early stopping in the training process. The training process was automatically terminated when the validation accuracy did not increase after 30 epochs, which took approximately 3 hours for the whole training process. We used the glorot\_uniform initialization and the Adam algorithm in keras. Segmentation runtime is 40-50 seconds for processing each set of testing data (size 256 ×128×256). Thus, our method is fast enough to be used in clinical practice.

\section{Discusion}
From Table 1 and Fig. 4, we can see that the result of CSF is worse than GM and WM due to the small number of CSF, so we can enhance the data of CSF to balance the samples in future research. 	

In the initial experiment, we tried normalization and Contrast-limited Adaptive Histogram Equalization (CLAHE) described in \cite{Chen2017VoxResNet}, but it benefited little for the segmentation result of ours method, so we did not use these preprocessing technology. On the other hand, we found BN layer had no effect on our model and we did not apply it either.\\	

To exploit the relationship between training sets with deep learning models, we used the method described in section 2.2 on the VoxResNet \cite{Chen2017VoxResNet} model. We found that specific training sets are needed for different deep learning models. Table 5 shows the most effective training sets found by the VoxResNet \cite{Chen2017VoxResNet} model. We can see that when we used the VoxResNet model, the performance would be better on using the whole training sets. In other words, each model has its own strong point and making full use of these strong point has great significance in improving the final segmentation performance, which is what we will continue to research in our future work.\\

\begin{table}
	\begin{tabular}{ccccc}
		\toprule
		Training set& Dice of CSF& Dice of GM&Dice of WM&Average dice\\
		\midrule
		1& 0.7555& 0.8861&0.8837&0.8418\\
		2& 0.7939& 0.8873&0.8868&0.8560\\
		3& 0.7586& 0.9166&0.9042&0.8598\\
		4& 0.7824& 0.8915&0.8898&0.8545\\
		\textbf{5}& \textbf{0.7983}& \textbf{0.9114}&\textbf{0.8980}&\textbf{0.8692}\\
		\bottomrule
		
	\end{tabular}
	\caption{The result of different training sets used on VoxResNet (training set 1 including subject 1,2,4,5,14;training set 2 including subject 1,2,3,4,5;training set 3 including subject10,11,12,13,14; training set 4 including subject 3,10,11,12,13 and training set 5 including subject 1,2,3,4,5,10,11,12,13,14. Testing set are subject 6 7 8 9 15 16 17 18)}
\end{table}

\section{Conclusions}
In this paper, we segmented MR brain tissue with modified U-shape model and achieved better result compared with some other method and our method can be used in clinical. In addition, we chose more valuable data based on modified U-Net from training sets. Finally, we proposed an effective strategy of MR brain tissue images’ suggestive annotation with trained model.\\

\bibliographystyle{IEEEtranS}
\bibliography{3D-DRL}

\end{document}